\documentclass[english,onecolumn,smallextended,final]{svjour3}
\usepackage[T1]{fontenc}
\usepackage[latin9]{inputenc}
\usepackage{textcomp}
\usepackage{url}
\usepackage{amsmath}
\usepackage{amssymb}
\usepackage{graphicx}
\usepackage{epstopdf}
\usepackage[numbers,sort]{natbib}
\makeatletter

\providecommand{\tabularnewline}{\\}

\makeatother

\usepackage{babel}
\begin{document}

\title{A Review of Learning Vector Quantization Classifiers}

\titlerunning{A Review of Learning Vector Quantization Classifiers}

\author{David Nova \and Pablo A. Estévez}

\authorrunning{David Nova \and Pablo A. Estévez}

\institute{D. Nova \at Department of Electrical Engineering, Faculty of Physical
and Mathematical Sciences, University of Chile, Santiago, Chile\\
\email{dnovai@ug.uchile.cl}\\
\and P. A. Estévez \at Department of Electrical Engineering and
Advanced Mining Technology Center, Faculty of Physical and Mathematical
Sciences, University of Chile, Santiago, Chile\\
\email{pestevez@ing.uchile.cl}}
\maketitle
\begin{abstract}
In this work we present a review of the state of the art of Learning
Vector Quantization (LVQ) classifiers. A taxonomy is proposed which
integrates the most relevant LVQ approaches to date. The main concepts
associated with modern LVQ approaches are defined. A comparison is
made among eleven LVQ classifiers using one real-world and two artificial
datasets. 
\end{abstract}

\section{Introduction}

Learning Vector Quantization (LVQ) is a family of algorithms for statistical
pattern classification, which aims at learning prototypes (codebook
vectors) representing class regions. The class regions are defined
by hyperplanes between prototypes, yielding Voronoi partitions. In
the late 80's Teuvo Kohonen introduced the algorithm LVQ1 \cite{Kohonen1997,Kohonen1988introduction},
and over the years produced several variants. Since their inception
LVQ algorithms have been researched by a small but active community.
A search on the ISI Web of Science in November, 2013, found 665 journal
articles with the keywords \textquotedblleft{}Learning Vector Quantization\textquotedblright{}
or \textquotedblleft{}LVQ\textquotedblright{} in their titles or abstracts.
This paper is a review of the progress made in the field during the
last 25 years.

LVQ algorithms are related to other competitive learning algorithms
such as self-organizing maps (SOMs) \cite{Kohonen1997} and c-means.
Competitive learning algorithms are based on the winner-take-all learning
rule, and variants in which only certain elements or neighborhoods
are updated during learning. The original LVQ algorithms and most
modern extensions use supervised learning for obtaining class-labeled
prototypes (classifiers). However, LVQ can also be trained without
labels by unsupervised learning for clustering purposes \cite{Bezdek1995two,Gonzalez1995analysis,Karayiannis1996fuzzy,Karayiannis1997methodology,Karayiannis1999axiomatic,Pal1993generalized}.
In this paper we will focus our review only on LVQ classifiers.

LVQ classifiers are particularly intuitive and simple to understand
because they are based on the notion of class representatives (prototypes)
and class regions usually in the input space (Voronoi partitions).
This is an advantage over multilayer perceptrons or support vector
machines (SVMs), which are considered to be black boxes. Moreover,
support vectors are extreme values (those having minimum margins)
of the datasets, while LVQ prototypes are typical vectors. Another
advantage of LVQ algorithms is that they are simple and fast, as a
result of being based on Hebbian learning. The computational cost
of LVQ algorithms depends on the number of prototypes, which are usually
a fixed number. SVMs depend on the number of training samples instead,
because the number of support vectors is a fraction of the size of
the training set. LVQ has been shown to be a valuable alternative
to SVMs \cite{Hofmann2012,Hammer2004relevance}.

LVQ classifiers try to approximate the theoretical Bayesian border,
and can deal directly with multi-class problems. The initial LVQ learning
rules were heuristic, and showed sensitivity to initialization, slow
convergence problems and instabilities. However, two main approaches
have been proposed defining explicit cost functions from which to
derive learning rules via steepest descent or ascent \cite{Sato1996,Seo2003,Seo2003soft},
and solving the problem of convergence of the original LVQ algorithms.
The first model is a generalization of LVQ called Generalized Learning
Vector Quantization (GLVQ) \cite{Sato1996}. In GLVQ a cost function
is defined in such a way that a learning rule is derived via the steepest
descent. This cost function has been shown to be related to a minimization
of errors, and a maximization of the margin of the classifier \cite{Biehl2007}.
The second approach is called Robust Soft-LVQ (RSLVQ) \cite{Seo2003,Seo2003soft},
in which a statistical objective function is used to derive a learning
rule by gradient ascent. The probability density function (pdf) of
the data is assumed to be a Gaussian mixture for each class. Given
a data point, the logarithm of the ratio of the pdf of the correct
class versus the pdf's of the incorrect classes serves as a cost function
to be maximized.

Other LVQ improvement deals with the initialization sensitivity of
the original LVQ algorithms and GLVQ \cite{Hammer2005supervised,Jirayusakul2007,Qin2004,Qin2005}.
Recent extensions of the LVQ family of algorithms substitute the Euclidean
distance with more general metric structures such as: weighted Euclidean
metrics \cite{Hammer2002}, adaptive relevance matrix metrics \cite{Schneider2009a},
pseudo-Euclidean metrics \cite{Hammer2011}, and similarity measures
in kernel feature space that lead to kernelized versions of LVQ \cite{Qin2004c}.

There are thousands of LVQ applications such as in: image and signal
processing \cite{xuan95a,tse95b,Blume1997,karayiannis00b,bashyal2008recognition,lendasse1998forecasting,nanopoulos2001feature},
the biomedical field and medicine \cite{pradhan1996detection,pesu1998classification,Anagnostopoulos01a,dieterle2003urinary,hung2011suppressed,dutta2011identification,savio2009classification,chen2012accelerometer},
and industry \cite{lieberman1997evaluation,jeng2000prediction,yang2001fuzzy,ahn2007intelligent,bassiuny2007fault,chang2007learning},
to name just a few. An extensive bibliography database is available
in \cite{centre2005}.

In this paper we present a comprehensive review of the most relevant
supervised LVQ algorithms developed since the original work of Teuvo
Kohonen. We introduce a taxonomy of LVQ classifiers, and describe
the main algorithms. We compare the performance of eleven LVQ algorithms
empirically on artificial and real-world datasets. We discuss the
advantages and limitations of each method depending on the nature
of the datasets.

The remainder of this paper is organized as follows: In section 2
a taxonomy of LVQ classifiers is presented. In section 3 the main
LVQ learning rules are described. In section 4 the results obtained
with eleven different LVQ methods in three different datasets are
shown. In section 5 some open problems are presented. Finally, in
section 6 conclusions are drawn.

\section{A Taxonomy of LVQ Classifiers}

\begin{figure}[t]
\includegraphics[width=1\textwidth]{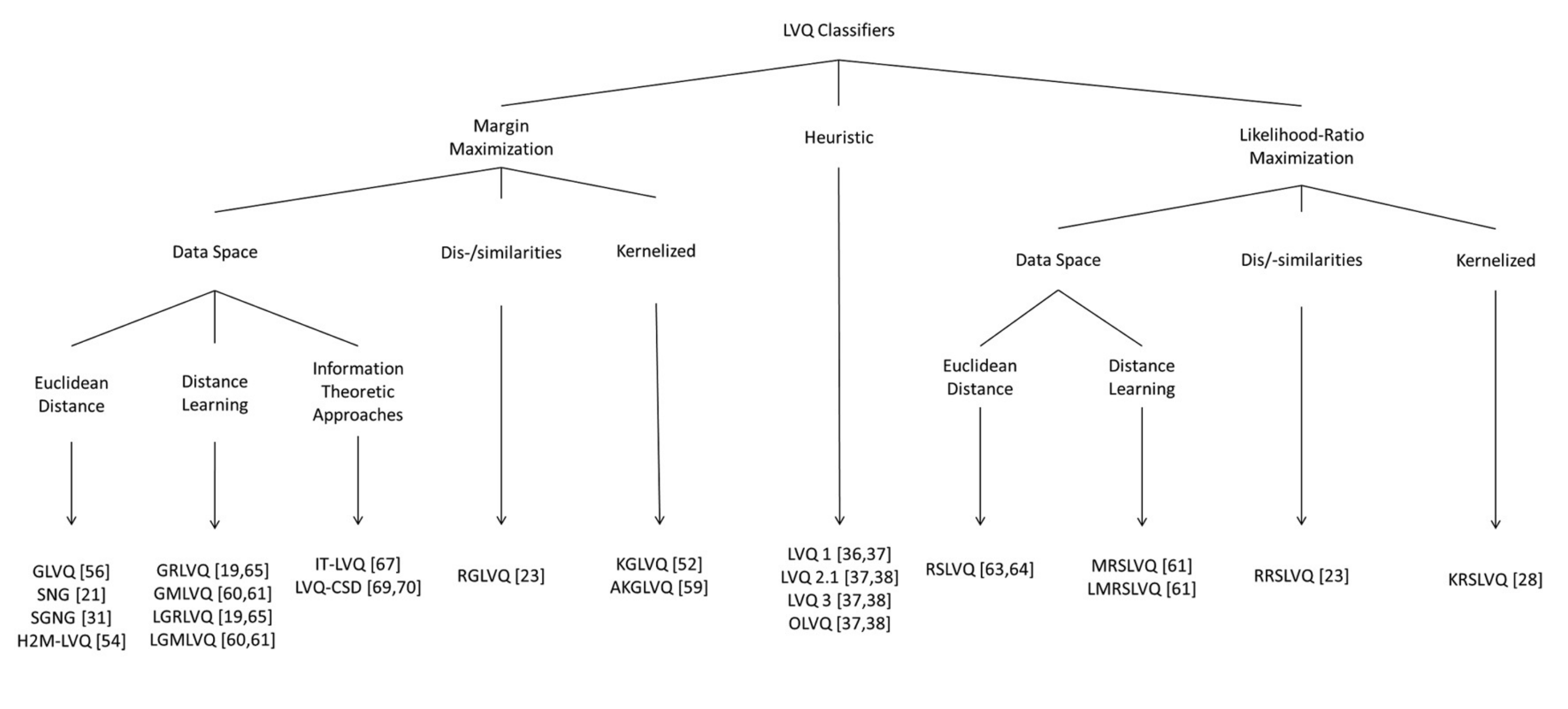}

\caption{Taxonomy of the most relevant Learning Vector Quantization classifiers
since the seminal work of Teuvo Kohonen in the late 80s.\label{fig:Taxonomy}}

\end{figure}

\subsection{Learning Vector Quantization}

Let $\mathbf{X}=\{(\mathbf{x}_{i},y_{i})\subset\mathbb{R}^{D}\times\{1,...,C\}|i=1,...,N\}$
be a training data set, where $\mathbf{x}=(x_{1},...,x_{D})\in\mathbb{R}^{D}$
are $D$-dimensional input samples, with cardinality $|\mathbf{X}|=N$;
$y_{i}\in\{1,...,C\}\; i=1,...,N$ are the sample labels, and $C$
is the number of classes. The neural network consists of a number
of prototypes, which are characterized by vectors $\mathbf{w}_{i}\in\mathbb{R}^{D}$,
for $i=1,...,M$, and their class labels $c(\mathbf{w}_{i})\in\{1,...,C\}$
with $\mathbf{Y}=\{c(\mathbf{w}_{j})\in\{1,...,C\}|j=1,...,M\}$.
The classification scheme is based on the best matching unit (BMU)
(winner-takes-all strategy). The receptive field of prototype $\mathbf{w}_{i}$
is defined as follows:
\begin{equation}
R^{i}=\{\mathbf{x}\in\mathbf{X}|\forall\mathbf{w}_{j}(j\neq i)\rightarrow d(\mathbf{w}_{i},\mathbf{x})\text{\ensuremath{\leq}}d(\mathbf{w}_{j},\mathbf{x})\},
\end{equation}
where $d(\mathbf{w},\mathbf{x})$ is a distance measure. Learning
aims at determining the weight vectors (prototypes), so that the training
data samples are mapped to their corresponding class labels.

Fig. \ref{fig:Taxonomy} shows a taxonomy of the most relevant LVQ
classifiers developed since the pioneer work of Kohonen. LVQ methods
are decomposed into 3 families: Kohonen's LVQ methods (middle branch:
heuristic), methods based on margin maximization (left branch), and
methods based on likelihood ratio maximization (right branch).

The original LVQ algorithm does not have an associated cost function
to ensure convergence. Some improvements such as LVQ2.1, LVQ3 or OLVQ
\cite{Kohonen1990} aim at achieving higher convergence speed or better
approximation of the Bayesian borders. All these LVQ versions are
based on Hebbian learning, i.e., are heuristic.

The original LVQ1 \cite{Kohonen1997,Kohonen1988introduction} corrected
only the winner prototype. This algorithm pulled the prototypes away
from the class borders. LVQ1 assumes a good initial state of the network,
i.e., it requires a preprocessing method. It also shows sensitivity
to overlapping data sets and in it, some neurons never learn the training
patterns. The LVQ2 algorithm updates two vectors at each step, the
winner and the runner-up. The purpose is to estimate differentially
the decision border towards the theoretical Bayes decision border.
But this algorithm makes corrections that are dependent on error only,
and present some instabilities. LVQ3 corrects the LVQ2 convergence
problem consisting of the location of prototypes changing in continued
learning by adding a stability factor.

\subsection{Margin Maximization}

In \cite{Crammer2002margin} a margin analysis of the original LVQ
algorithm was performed. There are two definitions of margin. The
first one is the sample-margin, which corresponds to the quantification
of samples which can travel through space without changing the classification
rate of the classifier. This is the definition used by SVMs. The second
definition is called the hypothesis-margin, which corresponds to the
quantification of the distance that the classifier (e.g. a hyperplane)
can be altered without changing the classification rate. This is the
definition used by Adaboost.

In the context of LVQ, the sample margin is hard to compute and numerically
unstable \cite{Crammer2002margin}. This is because small repositionings
of the prototypes might create large changes in the sample margin.
Crammer et al. \cite{Crammer2002margin} showed that the decision
borders of the original LVQ algorithms are hypothesis margin maximizers.
The margin is associated with generalization error bounds. Therefore,
maximizing the margin is equivalent to minimizing the generalization
error. Interestingly, the bound is dimension free, but depends on
the number of prototypes.

The left branch of our taxonomy shown in Fig. \ref{fig:Taxonomy}
corresponds to the LVQ methods based on a margin maximization approach.
The GLVQ \cite{Sato1996} proposed a cost function that aims at margin
maximization. This approach solves some limitations of the original
LVQ algorithms such as: slow convergence, initialization sensitivity,
limitations in multidimensional data where correlations exist between
dimensions, just to name a few.

The GLVQ cost function is defined as follows:
\begin{equation}
E_{GLVQ}=\sum_{i=1}^{N}\phi\left(\mu(\mathbf{x}_{i})\right),\label{eq:margin maximization cost function}
\end{equation}
where $\phi(\cdot)$ is the logistic sigmoid function, and $\mu$
is the relative distance difference
\begin{equation}
\mu(\mathbf{x}_{i},\mathbf{W})=\frac{d^{+}-d^{-}}{d^{+}+d^{-}},\label{eq:relative distance difference}
\end{equation}
 where $d^{+}=d(\mathbf{x}_{i},\mathbf{w}^{+})$ is the Euclidean
distance of data point $\mathbf{x}_{i}$ from its closest prototype
$\mathbf{w}^{+}$ having the same class label, and $d^{-}=d(\mathbf{x}_{i},\mathbf{w}^{-})$
is the Euclidean distance from the closest prototype $\mathbf{w}^{-}$
having a different class label. The term $d^{+}(\mathbf{x}_{i})-d^{-}(\mathbf{x}_{i})$
constitutes the hypothesis margin of an LVQ classifier according to
the winner-takes all rule \cite{Crammer2002margin,hammer2005generalization}.
Note that since GLVQ includes this margin in its cost function, it
can be described as a margin optimization learning algorithm. Generalization
bounds show that the larger the margin, the better the generalization
ability \cite{hammer2005generalization}. The cost function in Eq.
(\ref{eq:margin maximization cost function}) has been extended to
other distance metrics. In \cite{hammer2005generalization} a generalization
bound is derived for Generalized Relevance LVQ (GRLVQ), which uses
an adaptive metric.

\subsection{Likelihood Ratio Maximization}

In this section we describe in detail the cost function proposed by
Robust Soft-Learning Vector Quantization (RSLVQ) \cite{Seo2003,Seo2003soft}.
This cost function gives origin to the right branch of the taxonomy
illustrated in Fig. \ref{fig:Taxonomy}, which is based on a Gaussian
probabilistic model of data in forms of mixture models. It is assumed
that the probability density function (pdf) of the data is described
by a Gaussian mixture model for each class. The pdf of a sample that
is generated by the Gaussian mixture model of the correct class is
compared to the pdf of this sample that is generated by the Gaussian
mixture models of the incorrect classes. The logarithm of the ratio
between the correct mixture Gaussian model and the incorrect mixture
Gaussian models of probability densities is maximized. Let $\mathbf{W}$
be a set of labeled prototype vectors. The probability density of
the data is given by
\begin{equation}
p(\mathbf{x}|\mathbf{W})=\sum_{y=1}^{C}\sum_{\{j:c(\mathbf{w}_{j})=y\}}p(\mathbf{x}|j)P(j),
\end{equation}
where $C$ is the number of classes and $y$ is the class label of
the data points generated by component $j$. Also, $P(j)$ is the
probability that data points are generated by component $j$ of the
mixture and it can be chosen identically for each prototype $\mathbf{w}_{j}$.
$p(\mathbf{x}|j)$ is the conditional pdf that the component $j$
generates a particular data point $\mathbf{x}$ and it is a function
of prototype $\mathbf{w}_{j}$. The following likelihood ratio is
proposed as a cost function to be maximized:
\begin{equation}
E_{RSLVQ}=\sum_{i}^{N}\log\left(\frac{p(\mathbf{x}_{i},y|\mathbf{W})}{p(\mathbf{x}_{i}|\mathbf{W})}\right)\label{eq:likelihood ratio cost function}
\end{equation}
 where $p(\mathbf{x},y|\mathbf{W})$ is the pdf of a data point $\mathbf{x}$
that is generated by the mixture model for the correct class $y$,
and $p(\mathbf{x}|\mathbf{W})$ is the total probability density of
the data point $\mathbf{x}$. These probabilities are defined as follows:

\begin{eqnarray}
p(\mathbf{x}_{i},y_{i}|\mathbf{W}) & = & \sum_{(j:c(\mathbf{w}_{j})=y)}p(\mathbf{x}_{i}|j)P(j)\\
p(\mathbf{x}_{i}|\mathbf{W}) & = & \sum_{j}p(\mathbf{x}_{i}|j)P(j).
\end{eqnarray}

The conditional pdfs are assumed to be of the normalized exponential
form $p(\mathbf{x}|j)=K(j)\cdot\exp f(\mathbf{x},\mathbf{w}_{j},\sigma_{j}^{2})$,
with
\begin{equation}
f(\mathbf{x},\mathbf{w}_{j},\sigma\text{\texttwosuperior}_{j})=-\frac{d(\mathbf{x},\mathbf{w})}{2\sigma\text{\texttwosuperior}},\label{normalized exponential form RSLVQ}
\end{equation}
where $d(\mathbf{x},\mathbf{w})$ is the Euclidean distance measure.
Note that Eq. (\ref{normalized exponential form RSLVQ}) provides
a way to extend LVQ to other distance metrics by changing the distance
measure $d$.

\subsection{Distance Learning}

\begin{figure}
\begin{centering}
\begin{tabular}{ccc}
\includegraphics[scale=0.3]{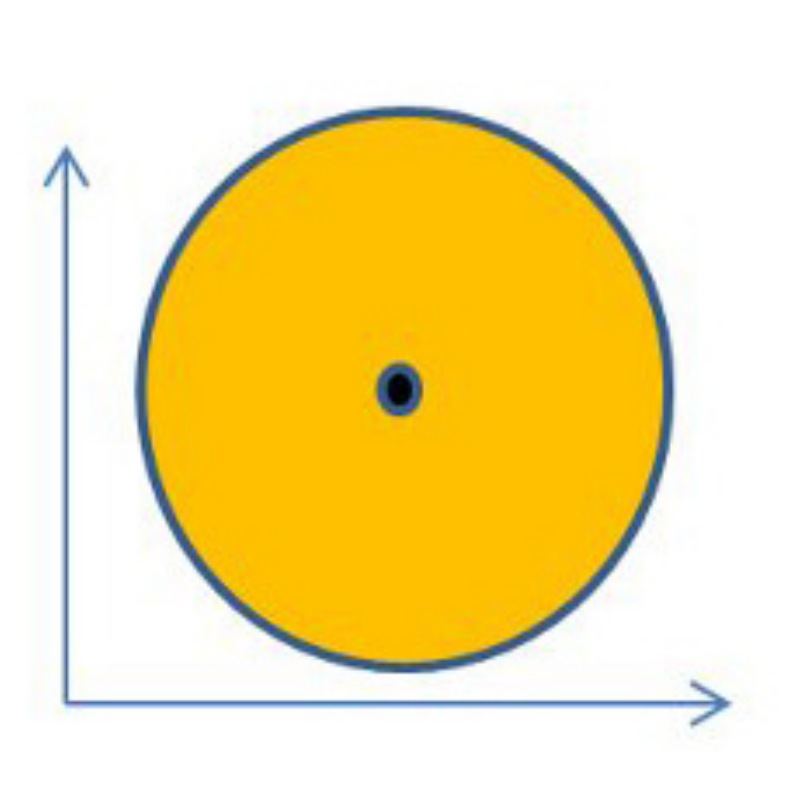} & \includegraphics[scale=0.3]{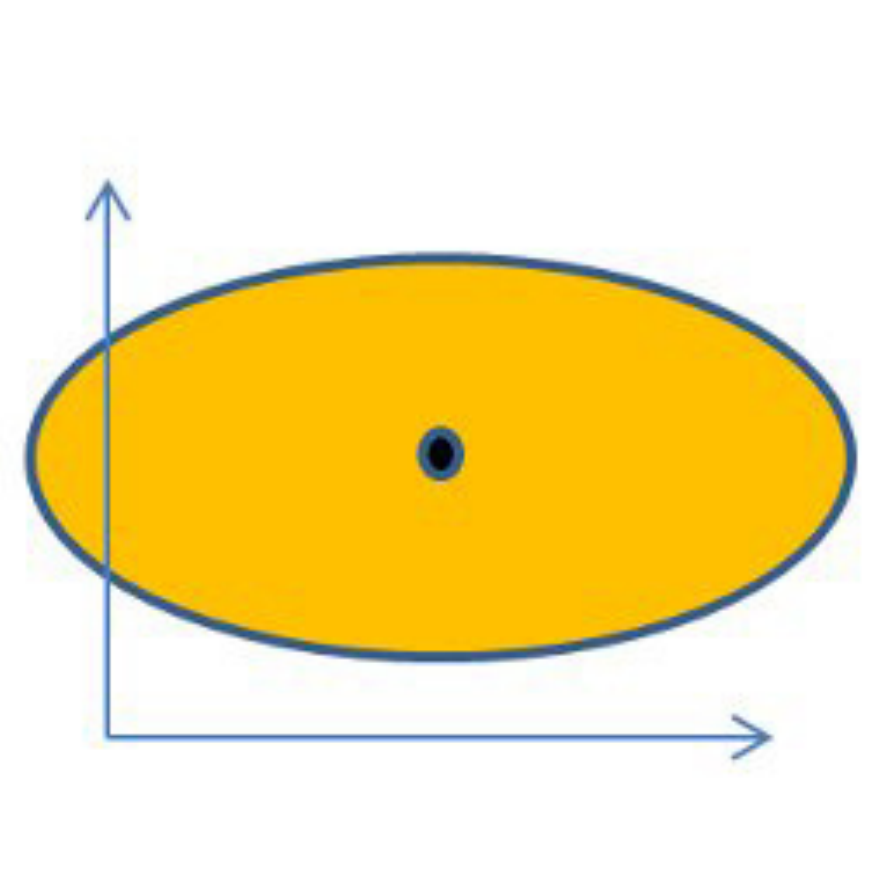} & \includegraphics[scale=0.3]{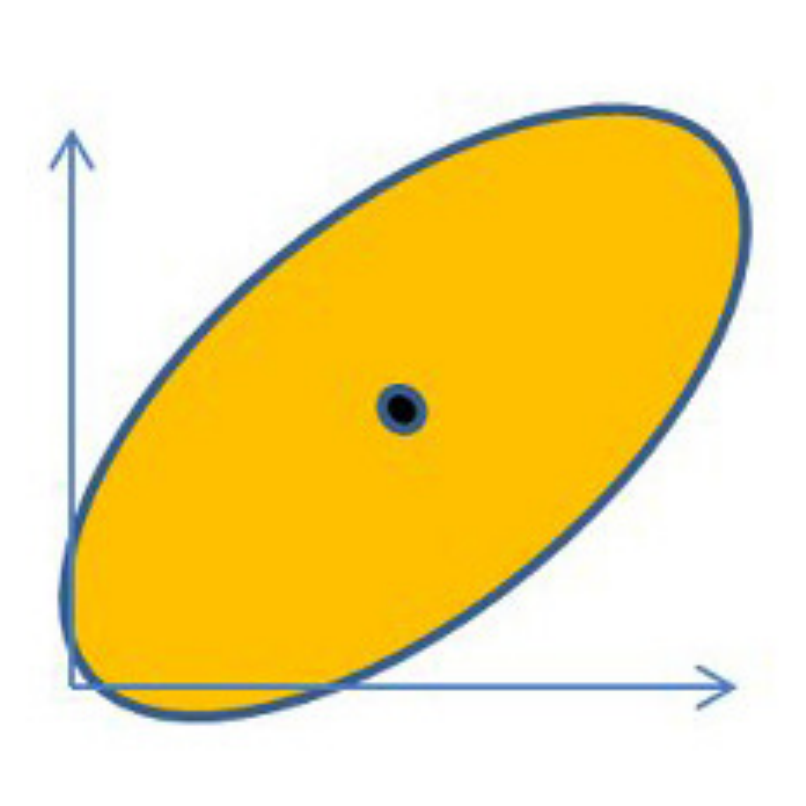}\tabularnewline
(a) & (b) & (c)\tabularnewline
\end{tabular}
\par\end{centering}

\caption{Shape of the receptive field of prototypes depending on the distance
measure. (a) Receptive field using the Euclidean distance measure.
(b) Receptive field using a diagonal matrix of relevance. (c) Receptive
field using a full matrix of relevance.\label{fig:Shape-of-receptiveFields}}
\end{figure}

The original LVQ, GLVQ and RSLVQ methods rely on the Euclidean distance.
This choice of metrics assumes a spherical receptive field of prototypes,
as shown in Fig. \ref{fig:Shape-of-receptiveFields}a. However, for
heterogeneous datasets, there can be different scaling and correlations
of the dimensions; and for high-dimensional datasets, the estimate
errors accumulate and can disrupt the classification \cite{Schneider2009a}.
In distance learning the distance measure is adaptive during training,
allowing us to get receptive fields for prototypes such as those shown
in Figs. \ref{fig:Shape-of-receptiveFields}b and \ref{fig:Shape-of-receptiveFields}c.
In the generalized matrix LVQ (GMLVQ) \cite{Schneider2009a,Schneider2009b}
a generalized distance metric is proposed as
\begin{equation}
d^{\Lambda}(\mathbf{w},\mathbf{x})=(\mathbf{x}-\mathbf{w})^{T}\Lambda(\mathbf{x}-\mathbf{w}),\label{eq:matrix_distance-1}
\end{equation}
 where $\Lambda$ is a full $D\times D$ matrix. To get a valid metric
$\Lambda$ must be positive (semi-) definite. This is achieved by
substituting
\begin{equation}
\Lambda=\Omega^{T}\Omega,
\end{equation}
which yields $\mathbf{u}^{T}\Lambda\mathbf{u}=\mathbf{u}^{T}\Omega^{T}\Omega\mathbf{u}=\left(\Omega^{T}\mathbf{u}\right)^{2}\geq0$
for all $\mathbf{u}$, where $\Omega\in\mathbb{R}^{D\times D}$. The
receptive field of prototype $\mathbf{w}_{i}$ becomes 
\begin{equation}
R_{\Lambda}^{i}=\{\mathbf{x}\in X|\forall\mathbf{w}_{j}(j\neq i)\rightarrow d^{\Lambda}(\mathbf{w_{i}},\mathbf{x})\leq d^{\Lambda}(\mathbf{w_{j}},\mathbf{x})\}.\label{eq:marix_repective_field}
\end{equation}
In this way arbitrary Euclidean metrics can be realized, in particular
correlations of dimensions and rotations of axes can be taken into
account. If $\Lambda$ is restricted to being diagonal then Eq. (\ref{eq:matrix_distance-1})
is reduced to 
\begin{equation}
d^{\lambda}(\mathbf{x},\mathbf{w})=\text{\ensuremath{\left\Vert \mathbf{x}-\mathbf{w}\right\Vert }}_{\lambda}^{2}=\sum_{j=1}^{D}\lambda_{j}(x_{j}-w_{j})^{2}.\label{eq:relevance distance}
\end{equation}
This simplification gives origin to what are called Generalized Relevance
approaches to LVQ (GRLVQ)\cite{Schneider2009b,Hammer2002,Strickert2001a}.

\subsection{Kernelization}

LVQ classifiers have been kernelized \cite{Qin2004c,Schleif2011a}.
A mapping function $\mathbf{\Phi}(\cdot)$ is defined in order to
realize a nonlinear transformation from the data space $\mathbb{R}^{D}$
to a higher dimensional possibly linearly separable feature space
$\mathbf{F}$, such as follows
\begin{equation}
\mathbf{\Phi}:\mathbb{R}^{D}\rightarrow\mathbf{F},\ \mathbf{x}\rightarrow\Phi(\mathbf{x}).
\end{equation}

A kernel function \cite{Scholkopf1999input} , can be represented
as a dot product and is usually chosen as a Gaussian kernel, 
\begin{equation}
k(\mathbf{x}_{i},\mathbf{x}_{j})=\mathbf{\Phi}(\mathbf{x}_{i})\cdot\mathbf{\Phi}(\mathbf{x}_{j})=\exp\left(-\frac{\left\Vert \mathbf{x}_{i}-\mathbf{x}_{j}\right\Vert ^{2}}{2\sigma\text{\texttwosuperior}}\right).
\end{equation}

The LVQ classifiers can then be applied in feature space, where the
prototypes are represented by an implicit form
\begin{equation}
\mathbf{w}_{j}^{F}=\sum_{m=1}^{N}\gamma_{jm}\mathbf{\Phi}(\mathbf{x}_{m}),\label{eq:prototype_kernel}
\end{equation}
where $\gamma_{j}\in\mathbb{R}^{N}$ are the combinatorial coefficient
vectors. The distance in feature space between an image $\mathbf{\Phi}(\mathbf{x}_{i})$
and a prototype vector $\mathbf{w}_{j}^{F}$ can be directly computed
using kernels:
\begin{eqnarray}
d^{F}(\mathbf{x}_{i},\mathbf{w}_{j}) & = & \left\Vert \mathbf{\Phi}(\mathbf{x})-\mathbf{w}_{j}^{F}\right\Vert =\sqrt{\left\Vert \mathbf{\Phi}(\mathbf{x})-\sum_{m=1}^{N}\gamma_{jm}\mathbf{\Phi}(\mathbf{x}_{m})\right\Vert ^{2}}\nonumber \\
 & = & k(\mathbf{x}_{i},\mathbf{x}_{i})-2\sum_{m=1}^{N}\gamma_{jm}k(\mathbf{x}_{i},\mathbf{x}_{m})+\sum_{s,t=1}^{N}\gamma_{js}\gamma_{jt}k(\mathbf{x}_{s},\mathbf{x}_{t}).\label{eq:kernel distance measure}
\end{eqnarray}
This is another form of the kernel trick, in which there is no need
of knowing the non-linear mapping $\mathbf{\Phi}(\cdot)$. This approach
is called kernel GLVQ (KGLVQ) \cite{Qin2004c}.

\subsection{Dis-/similarities}

Some problems do not allow a vectorial representation, e.g. alignment
of symbolic strings. In these cases the data can be represented by
pairwise similarities $s_{ij}=s(\mathbf{x}_{i},\mathbf{x}_{j})$ or
dissimilarities $d_{ij}=d(\mathbf{x}_{i},\mathbf{x}_{j})$ and their
corresponding matrices \textbf{$S$} and \textbf{$D$}, respectively.
These matrices are symmetric, i.e. $S=S^{t}$ and $D=D^{t}$, with
zero diagonals. It is easy to turn similarities into dissimilarities
and vice-versa, as shown in \cite{Pkekalska2005dissimilarity}. This
corresponds to what is called 'relational data' representation. Data
represented by pairwise dissimilarities with the previously mentioned
restrictions can always be embedded in a pseudo-Euclidean space \cite{Hammer2011,Pkekalska2005dissimilarity,hammer2011relational}.
A pseudo-Euclidean space is a vector space equipped with a symmetric
bilinear form composed of two parts: An Euclidean one (positive eigenvalues),
and a correction (negative eigenvalues) \cite{Pkekalska2005dissimilarity}.
The bilinear form is expressed as

\begin{equation}
\left\langle \mathbf{x},\mathbf{y}\right\rangle _{p,q}=\mathbf{x}^{t}I_{p,q}\mathbf{y},
\end{equation}
where $I_{p,q}$ is a diagonal matrix with $p$ entries $1$ and $q$
entries $-1$. This pseudo-Euclidean space is characterized by the
signature $(p,q,N-p-q)$, where the first $p$ components are Euclidean,
the next $q$ components are non-Euclidean, and $N-p-q$ components
are zeros. The prototypes are assumed to be linear combinations of
data samples:
\begin{equation}
\mathbf{w}_{j}=\sum_{m}\alpha_{jm}\mathbf{x}_{m},\ \textnormal{with }\sum_{m}\alpha_{jm}=1,\label{eq:linear combination prototype relational}
\end{equation}
where $\alpha_{j}=(\alpha_{j1},...,\alpha_{jN})$ is the vector of
coefficients, that describes the prototype $\mathbf{w}_{j}$ implicitly.
Unlike kernel approaches, dis-/similarities approaches do not assume
that the data is Euclidean. The distances between all data points
and prototypes are computed based on pairwise data similarities as
follows:
\begin{equation}
d^{\mathbf{D}}(\mathbf{x}_{i},\mathbf{w}_{j})=\left\Vert \mathbf{x}_{i}-\mathbf{w}_{j}\right\Vert ^{2}=\left[\mathbf{D}\cdot\alpha_{j}\right]_{i}-\frac{1}{2}\alpha_{j}^{T}\mathbf{D}\alpha_{j}.\label{eq:dissimilarities distance measure}
\end{equation}

\section{LVQ Learning Rules}

\subsection{LVQ 2.1}

\begin{figure}[t]
\begin{centering}
\includegraphics[scale=0.4]{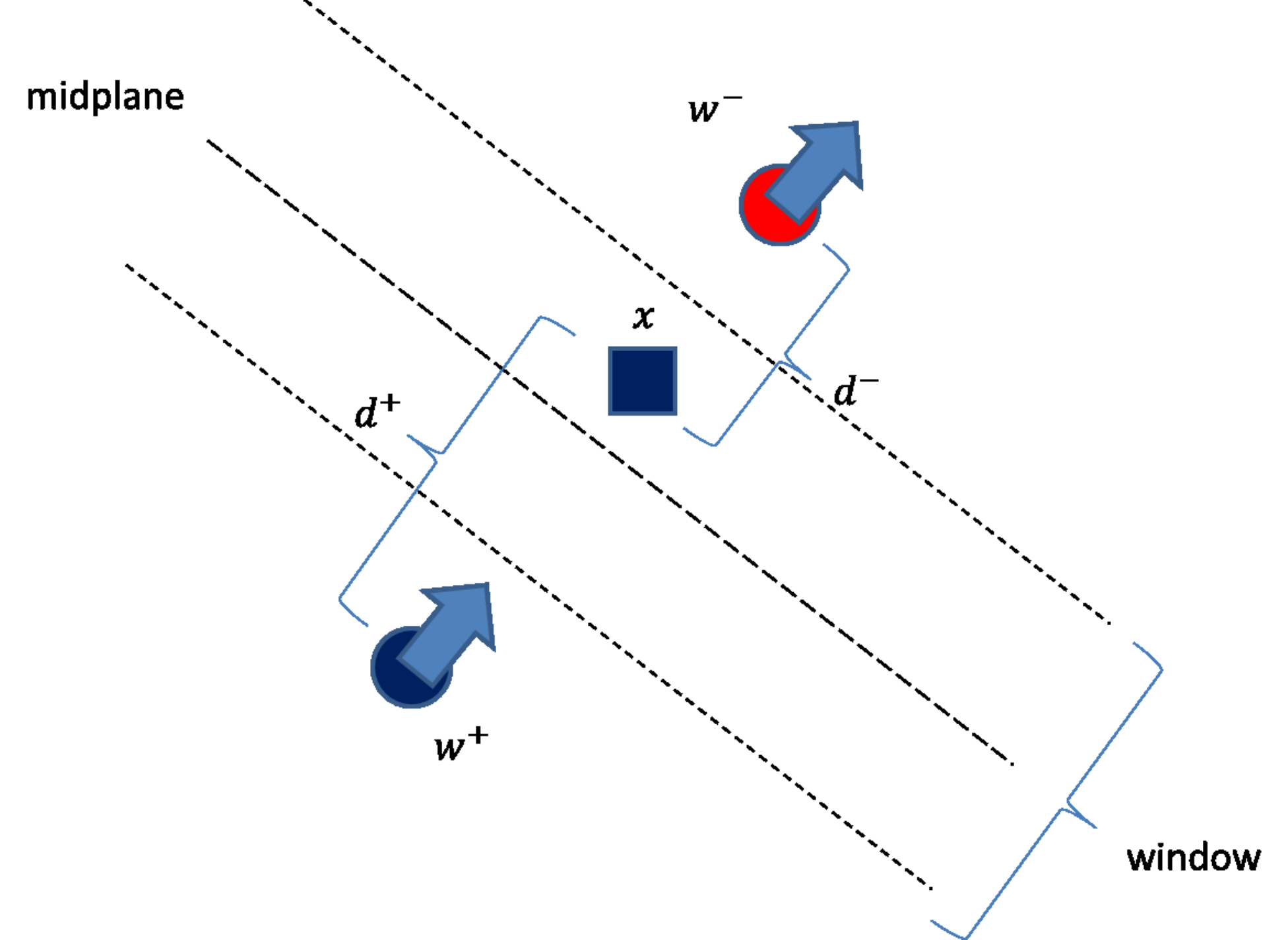}
\par\end{centering}

\caption{Schematic of the LVQ2.1 updating rule. The circles represent the prototypes;
the square represents an input sample; and the colors indicate different
class labels. In this example, the sample is incorrectly classified
and the nearest prototype of the correct class moves towards the sample,
while the nearest prototype of the incorrect class moves away from
the sample.\label{fig:Scheme-Window}}

\end{figure}

Among the initial variants proposed by Kohonen, the most popular is
LVQ2.1 \cite{Kohonen1990}, which is described in detail below. As
in LVQ2, two prototypes are updated at each step, the winner and the
runner-up. One of them, $\mathbf{w}^{+}$, belongs to the correct
class, while the other, $\mathbf{w}^{-}$, belongs to the incorrect
class. But now either the winner or the runner-up have the same class
label as the sample. Furthermore, the current input $\mathbf{x}_{i}$
must fall within a window defined around the mid-plane of vectors
$\mathbf{w}^{+}$and $\mathbf{w}^{-}$ (see Fig. \ref{fig:Scheme-Window}).
The updating rule is as follows:
\begin{eqnarray}
\mathbf{w}^{+}(t+1) & = & \mathbf{w}^{+}(t)+\epsilon(t)\cdot(\mathbf{x}_{i}-\mathbf{w}^{+}(t)),\textnormal{ if }c(\mathbf{w}^{+})=y\nonumber \\
\mathbf{w}^{-}(t+1) & = & \mathbf{w}^{-}(t)-\epsilon(t)\cdot(\mathbf{x}_{i}-\mathbf{w}^{-}(t)),\textnormal{ if }c(\mathbf{w}^{-})\neq y
\end{eqnarray}
where $\mathbf{w}^{+}$ ($\mathbf{w}^{-}$) is the closest prototype
to the input sample $\mathbf{x}_{i}$, with the same (different) class
label as the sample class label $y$, and $\epsilon\in]0,1[$ is the
learning rate. The prototypes, however, are changed only if the data
point $\mathbf{x}$ is close to the classification boundary, i.e.,
if it lands within a window of relative width $\omega$ defined by
\begin{equation}
\min\left(\frac{d(\mathbf{x},\mathbf{w}^{-})}{d(\mathbf{x},\mathbf{w}^{+})},\frac{d(\mathbf{x},\mathbf{w}^{+})}{d(\mathbf{x},\mathbf{w}^{-})}\right)<s,\; s=\frac{1-\omega}{1+\omega},
\end{equation}
where $d(\cdot)$ is the Euclidean distance measure and $\omega\in]0,1[$
(typically set to $\omega=0.2$ or $\omega=0.3$). A didactic scheme
of the LVQ2.1 update learning rule is illustrated in Fig. \ref{fig:Scheme-Window}.

\subsection{LVQ Methods Based on Margin Maximization}

\begin{figure}[t]
\begin{centering}
\begin{tabular}{cc}
\includegraphics[scale=0.6]{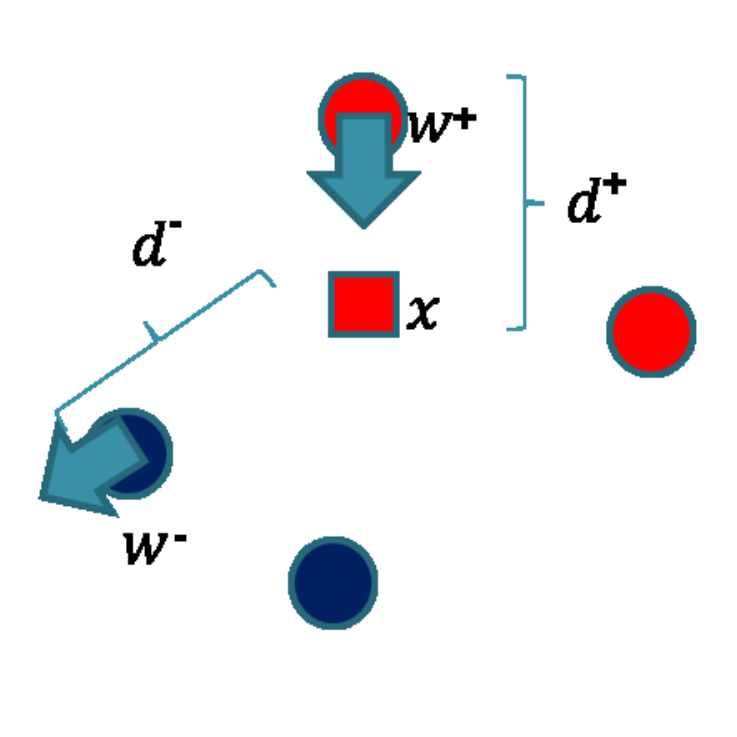} & \includegraphics[scale=0.6]{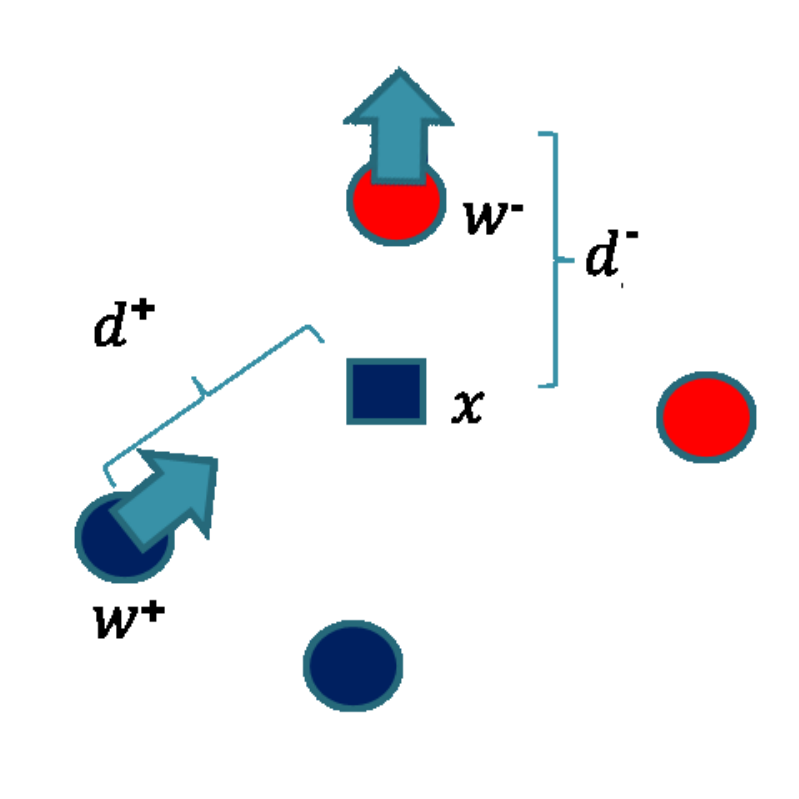}\tabularnewline
(a) & (b)\tabularnewline
\end{tabular}
\par\end{centering}

\caption{Schematic of the GLVQ learning rule which analyzes the behavior of
the two closest prototypes belonging to different classes. The circles
represent the prototypes; the square represents the input sample;
and the colors indicate the different class labels. (a) Rule behavior
when an input sample with a red class label is presented. (b) Rule
behavior when an input sample with a blue class label is presented.\label{fig:Scheme-GLVQ}}

\end{figure}

Using stochastic gradient descent to minimize Eq. (\ref{eq:margin maximization cost function}),
the following learning rules for GLVQ are obtained:

\begin{eqnarray}
\mathbf{w}^{+}(t+1) & = & \mathbf{w}^{+}(t)+2\cdot\epsilon\cdot\phi'(\mu(\mathbf{x}_{i}))\cdot\mu^{+}\cdot(\mathbf{x}_{i}-\mathbf{w}^{+})\nonumber \\
\mathbf{w}^{-}(t+1) & = & \mathbf{w}^{-}(t)-2\cdot\epsilon\cdot\phi'(\mu(\mathbf{x}_{i}))\cdot\mu^{-}\cdot(\mathbf{x}_{i}-\mathbf{w}^{-})\label{eq:glvq update rules}
\end{eqnarray}
where $\mu^{+}=\frac{2\cdot d^{-}}{(d^{-}+d^{+})^{2}}$ , $\mu^{-}=\frac{2\cdot d^{+}}{(d^{-}+d^{+})^{2}}$
and $\epsilon\in]0,1[$ is the learning rate. A didactic scheme of
the GLVQ learning rule is shown in Fig. \ref{fig:Scheme-GLVQ}.

Early LVQ versions assumed a fixed number of codebook vectors per
class and their initial values were set using an ad-hoc method. In
order to overcome these problems, alternative supervised versions
of Neural Gas (NG) \cite{martinetz1993neural} and Growing Neural
Gas (GNG) \cite{Fritzke1995} have been developed. These extensions
have been called Supervised Neural Gas (SNG) \cite{Hammer2005supervised}
and Supervised Growing Neural Gas (SGNG) \cite{Jirayusakul2007}. 

SNG adds neighborhood cooperativity to GLVQ solving the dependency
on the initialization. All prototypes $\mathbf{w}_{j}$ having the
same class as the current input sample $\mathbf{x}_{i}$ are adapted
towards the data point according to their ranking, and the closest
prototype of a different class label is moved away from the data sample.
SGNG incorporates plasticity by selecting an adaptive number of prototypes
during learning. Another method, the Harmonic to Minimum LVQ (H2MLVQ)
\cite{Qin2005} allows adapting all prototypes simultaneously in each
iteration. All prototypes having the same class label as the current
sample are updated. Likewise, all prototypes with a different label
from that of the sample are adjusted too, in which $d^{-}$ and $d^{+}$
in Eq. (\ref{eq:relative distance difference}) are replaced by harmonic
average distances \cite{Zhang1999k}. For more details the reader
is referred to \cite{Qin2004}. The previously mentioned methods tackle
the initialization sensitivity problem associated with the initial
position of the prototypes.

Other margin maximization methods are based on Information Theoretic
Learning (ITL) \cite{Torkkola2000,Torkkola2003,Lehn2005}. These methods
use a new divergence measure proposed by Principe et al. \cite{principe2000information},
based on the Cauchy-Schwarz inequality. Together with a consistently
chosen Parzen-estimator for the densities, it gives a numerically
well-behaved approach of information optimization prototype-based
vector quantization called Cauchy-Schwarz Divergence LVQ (CSDLVQ)\cite{Villmann2008,Villmann2011}.
These methods use fuzzy class labels to train the classifier.

\subsubsection{Distance Learning Approaches}

In the case of GRLVQ \cite{Hammer2002}, the relevance factors can
be determined by gradient descent, as follows:
\begin{equation}
\lambda_{m}(t+1)=\lambda_{m}(t)-\epsilon\phi'\left(\mu^{+}(x_{m}-w_{m}^{+})\text{\texttwosuperior}-\mu^{-}(x_{m}-w_{m}^{-})\text{\texttwosuperior}\right).
\end{equation}

The adaptation formulas are obtained by computing the derivatives
with respect to $\mathbf{w}$ and $\lambda$ using the following relative
distance difference:
\begin{equation}
\mu(\mathbf{x}_{i},\mathbf{W},\lambda)=\frac{d^{\lambda}(\mathbf{x}_{i},\mathbf{w}^{+})-d^{\lambda}(\mathbf{x}_{i},\mathbf{w}^{-})}{d^{\lambda}(\mathbf{x}_{i},\mathbf{w}^{+})+d^{\lambda}(\mathbf{x}_{i},\mathbf{w}^{-})},\label{eq:relative distance difference (relevance)}
\end{equation}
with $\mu^{+}=\frac{2d^{\lambda}(\mathbf{x}_{i},\mathbf{w}^{-})}{\left(d^{\lambda}(\mathbf{x}_{i},\mathbf{w}^{-})+d^{\lambda}(\mathbf{x}_{i},\mathbf{w}^{-})\right)^{2}}$
and $\mu^{-}=\frac{2d^{\lambda}(\mathbf{x}_{i},\mathbf{w}^{+})}{\left(d^{\lambda}(\mathbf{x}_{i},\mathbf{w}^{-})+d^{\lambda}(\mathbf{x}_{i},\mathbf{w}^{-})\right)^{2}}$
where $d^{\lambda}$ was defined in Eq. (\ref{eq:relevance distance}).
The updating rules for $\mathbf{w}$ can be obtained from Eq. (\ref{eq:glvq update rules})
by replacing $\mu^{+}$ and $\mu^{-}$ with those obtained from Eq.
(\ref{eq:relative distance difference (relevance)}) above.

Analogously, the relative distance difference for GMLVQ is as follows:
\begin{equation}
\mu(\mathbf{x}_{i},\mathbf{W},\Lambda)=\frac{d^{\Lambda}(\mathbf{x}_{i},\mathbf{w}^{+})-d^{\Lambda}(\mathbf{x}_{i},\mathbf{w}^{-})}{d^{\Lambda}(\mathbf{x}_{i},\mathbf{w}^{+})+d^{\Lambda}(\mathbf{x}_{i},\mathbf{w}^{-})}.
\end{equation}
 The update rules for GMLVQ are obtained by minimizing the cost function
defined in Eq. (\ref{eq:margin maximization cost function}) with
the generalized distance metric defined in Eq. (\ref{eq:matrix_distance-1}).
The update rules for $\mathbf{w}$ and $\Omega$ are as follows:
\begin{eqnarray}
\mathbf{w}^{+}(t+1) & = & \mathbf{w}^{+}(t)+\epsilon\cdot2\cdot\phi'\cdot\mu^{+}\cdot\Lambda\cdot(\mathbf{x}-\mathbf{w}^{+}(t))\nonumber \\
\mathbf{w}^{-}(t+1) & = & \mathbf{w}^{-}(t)-\epsilon\cdot2\cdot\phi'\cdot\mu^{-}\cdot\Lambda\cdot(\mathbf{x}-\mathbf{w}^{-}(t)),
\end{eqnarray}
and 
\begin{eqnarray}
\Omega_{lm}(t+1) & = & \Omega_{lm}(t)-\tilde{\epsilon}\cdot2\cdot\phi'\nonumber \\
 &  & \cdot(\mu\text{\textsuperscript{+}}\cdot((x_{m}-w_{m}^{+})[\Omega(\mathbf{x}-\mathbf{w}\text{\textsuperscript{+}})]_{l})\nonumber \\
 &  & -\mu\text{\textsuperscript{-}}\cdot((x_{m}-w_{m}^{-})[\Omega(\mathbf{x}-\mathbf{w}^{-})]_{l})),
\end{eqnarray}
where $l$ and $m$ specify matrix components, $\mu^{+}=\frac{2d^{\Lambda}(\mathbf{x}_{i},\mathbf{w}^{-})}{\left(d^{\Lambda}(\mathbf{x}_{i},\mathbf{w}^{-})+d^{\Lambda}(\mathbf{x}_{i},\mathbf{w}^{-})\right)^{2}}$
and $\mu^{-}=\frac{2d^{\Lambda}(\mathbf{x}_{i},\mathbf{w}^{+})}{\left(d^{\Lambda}(\mathbf{x}_{i},\mathbf{w}^{-})+d^{\Lambda}(\mathbf{x}_{i},\mathbf{w}^{-})\right)^{2}}$,
and the distance $d^{\Lambda}(\cdot)$ is described in Eq. (\ref{eq:matrix_distance-1}).

\subsubsection{Kernel Approaches}

An extension of GLVQ is the Kernel GLVQ (KGLVQ) \cite{Qin2004c} .
KGLVQ projects the data space non-linearly into a higher dimensional
feature space, and then applies the GLVQ algorithm in this newly formed
feature space. The prototypes are represented in an implicit form
as shown by Eq. (\ref{eq:prototype_kernel}). The updating rules of
the GLVQ algorithm in Eq. (\ref{eq:glvq update rules}) can be generalized
from the data space into the feature space F by using the following
relative distance difference:

\begin{equation}
\mu(\mathbf{\Phi}(\mathbf{x}_{i}),\mathbf{W}^{F})=\frac{d^{F}(\mathbf{\Phi}(\mathbf{x}_{i}),\mathbf{w}^{+})-d^{F}(\mathbf{\Phi}(\mathbf{x}_{i}),\mathbf{w}^{-})}{d^{F}(\mathbf{\Phi}(\mathbf{x}_{i}),\mathbf{w}^{+})+d^{F}(\mathbf{\Phi}(\mathbf{x}_{i}),\mathbf{w}^{-})},
\end{equation}
and its derivatives $\mu\text{\textsuperscript{+}}=\frac{2d^{F}(\mathbf{x}_{i},\mathbf{w}^{-})}{\left(d^{F}(\mathbf{x}_{i},\mathbf{w}^{+})+d^{F}(\mathbf{x}_{i},\mathbf{w}^{-})\right)^{2}}$
and $\mu^{-}=\frac{2d^{F}(\mathbf{x}_{i},\mathbf{w}^{+})}{\left(d^{F}(\mathbf{x}_{i},\mathbf{w}^{+})+d^{F}(\mathbf{x}_{i},\mathbf{w}^{-})\right)^{2}}$
where the metric $d^{F}$ is defined in Eq. (\ref{eq:kernel distance measure}).

The equivalent updating rules in terms of adjusting parameters $\gamma$
in Eq. (\ref{eq:prototype_kernel}) are:
\begin{eqnarray}
\gamma_{s}^{+}(t+1) & = & \begin{cases}
\left[1-\epsilon\cdot\phi'\cdot\mu^{+}\right]\cdot\gamma_{s}^{+}(t) & \textrm{if }\mathbf{x}_{s}\neq\mathbf{x}_{i}\\
\left[1-\epsilon\cdot\phi'\cdot\mu^{+}\right]\cdot\gamma_{s}^{+}(t)+\epsilon\cdot\phi'\cdot\mu^{+} & \textrm{if }\mathbf{x}_{s}=\mathbf{x}_{i}
\end{cases}\nonumber \\
\gamma_{s}^{-}(t+1) & = & \begin{cases}
\left[1+\epsilon\cdot\phi'\cdot\mu^{-}\right]\cdot\gamma_{s}^{-}(t) & \textrm{if }\mathbf{x}_{s}\neq\mathbf{x}_{i}\\
\left[1+\epsilon\cdot\phi'\cdot\mu^{-}\right]\cdot\gamma_{s}^{-}(t)-\epsilon\cdot\phi'\cdot\mu^{-} & \textrm{if }\mathbf{x}_{s}=\mathbf{x}_{i}
\end{cases},
\end{eqnarray}
where $s$ specifies combinatorial coefficient vectors of $\gamma$,
$\gamma^{+}$ is the coefficient vector associated with the nearest
prototype $\mathbf{w}^{+}$having the same class label as $\mathbf{x}_{i}$,
and $\gamma^{-}$ is the coefficient vector associated with the nearest
prototype $\mathbf{w}^{-}$ having a different class label from $\mathbf{x}_{i}$.

Furthermore, in \cite{Schleif2011a} KGLVQ is extended to get the
Nystrom-Approximation Generalized LVQ (AKGLVQ), where sparsity is
imposed and a Nystrom approximation technique is used to reduce the
learning complexity in large data sets.

\subsubsection{Dis-/similarities Approaches}

In \cite{Hammer2011} a new method was proposed using dis-/similarity
data by representing it in an implicit form. This approach is extended
to the margin maximization technique giving origin to the relational
GLVQ (RGLVQ).

An implicit representation of the prototypes is assumed as described
through a model by Eq. (\ref{eq:linear combination prototype relational}).
Once again the starting point is the GLVQ updating rule (Eq. (\ref{eq:glvq update rules}))
where the relative distance difference is now computed as

\begin{equation}
\mu(\mathbf{x}_{i},\mathbf{W},D)=\frac{d^{\mathbf{D}}(\mathbf{x}_{i},\mathbf{w}^{+})-d^{\mathbf{D}}(\mathbf{x}_{i},\mathbf{w}^{-})}{d^{\mathbf{D}}(\mathbf{x}_{i},\mathbf{w}^{+})+d^{\mathbf{D}}(\mathbf{x}_{i},\mathbf{w}^{-})},
\end{equation}
where $d^{\mathbf{D}}(\cdot)$ is the distance defined in Eq. (\ref{eq:dissimilarities distance measure}),
and $\mu\text{\textsuperscript{+}}=\frac{2d^{\mathbf{D}}(\mathbf{x}_{i},\mathbf{w}^{-})}{\left(d^{\mathbf{D}}(\mathbf{x}_{i},\mathbf{w}^{+})+d^{\mathbf{D}}(\mathbf{x}_{i},\mathbf{w}^{-})\right)^{2}}$
and $\mu^{-}=\frac{2d^{\mathbf{D}}(\mathbf{x}_{i},\mathbf{w}^{+})}{\left(d^{\mathbf{D}}(\mathbf{x}_{i},\mathbf{w}^{+})+d^{\mathbf{D}}(\mathbf{x}_{i},\mathbf{w}^{-})\right)^{2}}$
are the respective partial derivatives of the relative distance difference.
The $d_{ij}$ are dissimilarity matrix elements of $D\in\mathbb{R}^{N\times N}$;
$\alpha_{m}^{+}$ and $\alpha_{m}^{-}$ are the coefficient vector
elements associated with the nearest prototype $\mathbf{w}^{+}$ having
the same class label as $\mathbf{x}_{i}$, and the nearest prototype
$\mathbf{w}^{-}$ having a different class label from $\mathbf{x}_{i}$,
respectively. The equivalent updating rules for the $\alpha$ parameters
are: 
\begin{eqnarray}
\alpha_{m}^{+}(t+1) & = & \alpha_{m}^{+}(t)-\epsilon\cdot\phi'\cdot\mu^{+}\cdot(d_{im}-\sum_{l}d_{lm}\alpha_{l}^{+})\nonumber \\
\alpha_{m}^{+}(t+1) & = & \alpha_{m}^{-}(t)+\epsilon\cdot\phi'\cdot\mu^{-}\cdot(d_{im}-\sum_{l}d_{lm}\alpha_{l}^{-}).
\end{eqnarray}

\subsection{\emph{Methods Based on Likelihood Ratio Maximization}}

In RSLVQ \cite{Seo2003,Seo2003soft}, the updating rule for the prototypes
is derived from the likelihood ratio Eq. (\ref{eq:likelihood ratio cost function})
by maximizing this cost function through gradient ascent,

\begin{equation}
\mathbf{w}_{j}(t+1)=\mathbf{w}_{j}(t)+\epsilon(t)\frac{\partial}{\partial\mathbf{w}_{j}}\log\left(\frac{p(\mathbf{x}_{i},y_{i}|\mathbf{W})}{p(\mathbf{x}_{i}|\mathbf{W})}\right).\label{eq:rslvq_updaterule}
\end{equation}
Eq. (\ref{eq:rslvq_updaterule}) can be re-written as follows:

\begin{equation}
\mathbf{w}_{j}(t+1)=\mathbf{w}_{j}(t)+\frac{\epsilon}{\sigma^{2}}\left\lbrace \begin{array}{ccc}
(P_{y}(j|\mathbf{x})-P(j|\mathbf{x}))(\mathbf{x}-\mathbf{w}_{j}), &  & \textnormal{if }c(\mathbf{w}_{j})=y,\\
-P(j|\mathbf{x})(\mathbf{x}-\mathbf{w}_{j}), &  & \textnormal{if }c(\mathbf{w}_{j})\neq y,
\end{array}\right.\label{eq:RSLVQ_update_rule_B}
\end{equation}
where $\epsilon\in]0,1[$ is the learning rate, $y$ is the class
label of the data points generated by component $j$, and $P_{y}(j|\mathbf{x})$
and $P(j|\mathbf{x})$ are assignment probabilities described as follows:
\begin{equation}
P_{y}(j|\mathbf{x})=\frac{p(j)\exp\left(f\left(\mathbf{x},\mathbf{w}_{j},\sigma_{j}^{2}\right)\right)}{\sum_{\{i:c(\mathbf{w}_{j})=y\}}p(i)\exp\left(f\left(\mathbf{x},\mathbf{w}_{i},\sigma_{i}^{2}\right)\right)},\label{eq:probability P_y}
\end{equation}
\begin{equation}
P(j|\mathbf{x})=\frac{p(j)\exp\left(f\left(\mathbf{x},\mathbf{w}_{j},\sigma_{j}^{2}\right)\right)}{\sum_{i=1}^{M}p(i)\exp\left(f\left(\mathbf{x},\mathbf{w}_{i},\sigma_{i}^{2}\right)\right)}.\label{eq:Probability P}
\end{equation}

Note that Eq. (\ref{eq:RSLVQ_update_rule_B}) provides a way to extend
LVQ to other distance measures by changing $f(\mathbf{x},\mathbf{w},\sigma^{2})$
in Eqs. (\ref{eq:probability P_y}-\ref{eq:Probability P}).

\subsubsection{Distance Learning Approaches}

The matrix learning scheme was applied to RSLVQ \cite{Schneider2009a}
obtaining what was named the Matrix RSLVQ (MRSLVQ) and its local version,
Local MRSLVQ (LMRSLVQ). The latter consists of a local distance adaptation
for each prototype, which in these cases is a full matrix of relevance
factors. As it is based on RSLVQ, this approach uses the conditional
probability density function $p(\mathbf{x}|j)$ in which component
$j$ generates a particular data point $\mathbf{x}$. It is a function
of prototype $\mathbf{w}_{j}$ and it is assumed to have a normalized
exponential form $p(\mathbf{x}|j)=K(j)\cdot\exp f(\mathbf{x},\mathbf{w}_{j},\sigma_{j}^{2})$.
In matrix learning Eq. (\ref{normalized exponential form RSLVQ})
is substituted with Eq. (\ref{eq:normalized exponential form MRSLVQ})
where distance $d^{\Lambda}$, defined in Eq. (\ref{eq:matrix_distance-1}),
is used: 
\begin{equation}
f(\mathbf{x},\mathbf{w}_{j},\sigma_{j}^{2},\Lambda)=-\frac{d^{\Lambda}(\mathbf{x},\mathbf{w})}{2\sigma^{2}}.\label{eq:normalized exponential form MRSLVQ}
\end{equation}
The updating rules for $\mathbf{w}$ and $\Omega$ are obtained in
a way similar to RSLVQ but using the adaptive distance $d^{\Lambda}$,
as follows:
\begin{equation}
\mathbf{w}_{j}(t+1)=\mathbf{w}_{j}(t)+\frac{\epsilon}{\sigma^{2}}\begin{cases}
(P_{y}(j|\mathbf{x})-P(j|\mathbf{x}))\Lambda(\mathbf{x}-\mathbf{w}_{j}), & \textnormal{if }c(\mathbf{w}_{j})=y\\
-P(j|\mathbf{x})\Lambda(\mathbf{x}-\mathbf{w}_{j}), & \textnormal{if }c(\mathbf{w}_{j})\neq y
\end{cases},
\end{equation}
where $y$ is the class label of the data points generated by component
$j$, and
\begin{eqnarray}
\Omega_{lm}(t+1) & = & \Omega_{lm}(t)-\frac{\tilde{\epsilon}}{\sigma^{2}}\cdot\nonumber \\
 &  & \sum[(\delta_{y,l_{i}}(P_{y}(j|\mathbf{x})-P(j|\mathbf{x}))-(1-\delta_{y,l_{i}})P(j|\mathbf{x}))\nonumber \\
 &  & \cdot([\Omega(\mathbf{x}-\mathbf{w}_{j})]_{l}(x_{m}-w_{j,m}))],
\end{eqnarray}
where $l$ and $m$ are matrix elements. This approach can be extended
to the local matrix for every prototype as proposed in \cite{Schneider2009a}.

\subsubsection{Kernel Approaches}

LVQ methods based on likelihood ratio maximization have been extended
to feature space by using kernels. A kernelized version of RSLVQ was
proposed in \cite{Hofmann2012} and called Kernel Robust Soft Learning
Vector Quantization (KRSLVQ). The prototypes are represented as a
linear combination of data images in the feature space as described
in Eq. (\ref{eq:prototype_kernel}). As in RSLVQ, a Gaussian mixture
model is assumed, but the Euclidean distance is measured in feature
space, using the distance $d^{F}$ defined in Eq. (\ref{eq:kernel distance measure}).
Consequently Eq. (\ref{normalized exponential form RSLVQ}) is substituted
with
\begin{equation}
f(\Phi(\mathbf{x}),\mathbf{w}_{j}^{F},\sigma_{j}^{2})=-\frac{d^{F}\left(\Phi(\mathbf{x}),\mathbf{w}_{j}^{F}\right)}{2\sigma\text{\texttwosuperior}}.\label{eq:exponential form KRSLVQ}
\end{equation}

The updating rules of KRSLVQ are
\begin{equation}
\mathbf{w}_{j}(t+1)=\mathbf{w}_{j}(t)+\frac{\epsilon}{\sigma^{2}}\begin{cases}
-(P_{y}(j|\Phi(\mathbf{x}_{i}))-P(j|\Phi(\mathbf{x}_{i})))\gamma_{jm} & \textrm{if }\mathbf{x}_{m}\neq\mathbf{x}_{i},\ c(\mathbf{w}_{j})=y\\
(P_{y}(j|\Phi(\mathbf{x}_{i}))-P(j|\Phi(\mathbf{x}_{i})))(1-\gamma_{jm}) & \textrm{if }\mathbf{x}_{m}=\mathbf{x}_{i},\ c(\mathbf{w}_{j})=y\\
P(j|\Phi(\mathbf{x}_{i}))\gamma_{jm} & \textrm{if }\mathbf{x}_{m}\neq\mathbf{x}_{i},\ c(\mathbf{w}_{j})\neq y\\
-P(j|\Phi(\mathbf{x}_{i}))(1-\gamma_{jm}) & \textrm{if }\mathbf{x}_{m}=\mathbf{x}_{i},\ c(\mathbf{w}_{j})\neq y
\end{cases}
\end{equation}

\subsubsection{Dis-/similarities Approaches}

Dis-/similarities approaches allow extending the likelihood-ratio
maximization LVQ methods to a pseudo-Euclidean space giving origin
to relational RSLVQ \cite{hammer2011relational}. The prototypes are
represented in an implicit form by a linear combination of the data
points as in Eq. (\ref{eq:linear combination prototype relational}).
Remember that $D$ represents a dissimilarity matrix which is symmetric
and with diagonal elements $d_{ii}=0$. The dissimilarities are computed
by using Eq. (\ref{eq:dissimilarities distance measure}). In this
approach the argument of the normalized exponential takes the following
form:
\begin{equation}
f(\mathbf{x},\mathbf{w}_{j},\sigma_{j}^{2})=-\frac{d\left(\mathbf{x},\mathbf{w}_{j}\right)}{2\sigma\text{\texttwosuperior}}=-\frac{\left[D\cdot\alpha_{j}\right]-\frac{1}{2}\cdot\alpha_{j}^{t}D\text{\ensuremath{\alpha}}_{j}}{2\sigma^{2}}.\label{eq:exponential form RRSLVQ}
\end{equation}

Using stochastic gradient ascent the following updating rules are
obtained:
\begin{equation}
\alpha_{jm}(t+1)=\alpha_{jm}(t)-\epsilon\frac{1}{2\sigma^{2}}\begin{cases}
\left[\frac{p(\mathbf{x}|j)}{\sum_{j:c(\mathbf{w}_{j})=y}p(\mathbf{x}_{i}|j)}-\frac{p(\mathbf{x}_{i}|j)}{\sum_{j}p(\mathbf{x}_{i}|j)}\right]\\
\cdot\left[d_{im}-\sum_{l}d_{lm}\alpha_{jm}\right] & \textrm{if }c(\mathbf{w}_{j})=y\\
\left[\frac{p(\mathbf{x}_{i}|j)}{\sum_{j:c(\mathbf{w}{}_{j})=y}p(\mathbf{x}_{i}|j)}\right]\\
\cdot\left[d_{im}-\sum_{l}d_{lm}\alpha_{jm}\right] & \textrm{if }c(\mathbf{w}_{j})\neq y
\end{cases},
\end{equation}
where $p(\mathbf{x}_{i}|j)=K\cdot\exp\left(-\frac{\left[D\cdot\alpha_{j}\right]-\frac{1}{2}\cdot\alpha_{j}^{t}D\text{\ensuremath{\alpha}}_{j}}{2\sigma^{2}}\right)$
and $\alpha_{j}$ is a coefficient vector which implicitly describes
$\mathbf{w}_{j}$ through Eq. (\ref{eq:linear combination prototype relational}).

\section{Results}

In this section the results of three different experiments are presented.
The first one employs the Multi-modal dataset \cite{Qin2005}, which
is used for studying the sensitivity to the initialization or initial
position of the prototypes. The second dataset is Image Segmentation
\cite{Frank2010}, which is used for comparing the performance of
the different methods when the nature of the features in the data
is heterogeneous. Finally, the third dataset is USPS \cite{trevor2001elements},
which allows us to compare the performance of the LVQ classifiers
of the different methods in a real-world problem.

The multi-modal dataset \cite{Qin2005} has three classes C1, C2 and
C3, with 1200 training samples per class. The training samples in
class C3 are distributed in three clusters, while those in classes
C1 and C2 have multi-modal distributions. Class C1 consists of 15
sub-clusters and the number of samples per cluster are 50, 50, 50,
50, 50, 50, 50, 50, 50, 150, 150, 150, 100, 100 and 100, respectively.
Class C2 is composed of 12 sub-clusters and the number of samples
per cluster are 100, 100, 100, 50, 50, 50, 50, 50, 50, 200, 200 and
200, respectively.

The USPS dataset consists of 9298 images of handwritten digits 0-9
(10 classes) of 16x16 pixels in gray scale which are split into 7291
training set images and 2007 test set images \cite{trevor2001elements}.
In this experiment we used the original dataset and also a subset
of 2000 images, which is named USPS{*}. This smaller dataset is used
for comparison purposes with other works published in the literature.

The Image Segmentation dataset consists of 2100 samples having 12
features which correspond to 3x3 pixel regions extracted from outdoor
images. There are 7 classes which are: brick-face, sky, foliage, cement,
window, path and grass \cite{Frank2010}. Because features 3-5 are
constants, they were eliminated.

10-fold cross validation was used for comparing the performance of
the different LVQ algorithms. In addition, a multi-comparison statistical
test was used to compare the means of all pairs of LVQ classifiers.
This test involves comparing many group means, i.e. pairs of simple
t-test comparisons are realized and then a Bonferroni adjustment is
done to compensate the critical value used in a multiple comparison
procedure \cite{Hochberg1987,salzberg1997comparing}.

For all LVQ classifiers the following learning rate was used:

\begin{equation}
\epsilon(t)=\frac{\epsilon_{0}}{(1+\tau\cdot(t-t_{0}))},
\end{equation}
where $\epsilon_{0}$ is the initial value of the learning rate, $\tau=0.0001$
for Multi-modal, $\tau=0.001$ for the Image Segmentation dataset,
and $\tau=0.001$ for USPS; and $t_{0}$ is the start time for the
learning rate with $T_{max}=2000$ as the maximum number of training
epochs for all experiments.

Table 1 shows a summary of 11 LVQ classifiers chosen for comparison.
All of them are compared using the 3 datasets described above.

\begin{table}[t]
\caption{Summary of 11 LVQ Classifiers}

\centering{}{\scriptsize }%
\begin{tabular}{lllll}
{\scriptsize Name} & {\scriptsize Characteristics} & {\scriptsize Parameters} & {\scriptsize Distance} & {\scriptsize Constraint}\tabularnewline
\hline 
\hline 
{\scriptsize LVQ 2.1} & {\scriptsize Heuristic} & {\scriptsize $\left\{ \epsilon,\omega,\mathbf{W}\right\} $} & {\scriptsize Euclidean distance } & {\scriptsize $\min\left(\frac{d(\mathbf{x},\mathbf{w}^{-})}{d(\mathbf{x},\mathbf{w}^{+})},\frac{d(\mathbf{x},\mathbf{w}^{+})}{d(\mathbf{x},\mathbf{w}^{-})}\right)<s,$}\tabularnewline
 &  &  & {\scriptsize $d(\mathbf{x},\mathbf{w})=\left\Vert \mathbf{x}-\mathbf{w}\right\Vert $} & {\scriptsize $s=\frac{1-\omega}{1+\omega}$}\tabularnewline
\hline 
{\scriptsize GLVQ} & {\scriptsize Margin Maximization} & {\scriptsize $\left\{ \epsilon,\mathbf{W}\right\} $} & {\scriptsize Euclidean distance} & {\scriptsize no }\tabularnewline
 &  &  & {\scriptsize $d(\mathbf{x},\mathbf{w})=\left\Vert \mathbf{x}-\mathbf{w}\right\Vert $} & \tabularnewline
\hline 
{\scriptsize RSLVQ} & {\scriptsize Likelihood Ratio Maximization} & {\scriptsize $\left\{ \epsilon,\sigma,\mathbf{W}\right\} $} & {\scriptsize Euclidean Distance} & {\scriptsize no }\tabularnewline
 &  &  & {\scriptsize $d(\mathbf{x},\mathbf{w})=\left\Vert \mathbf{x}-\mathbf{w}\right\Vert $} & \tabularnewline
\hline 
{\scriptsize SNG} & {\scriptsize Margin Maximization} & {\scriptsize $\left\{ \epsilon,\mathbf{W}\right\} $} & {\scriptsize Euclidean Distance} & {\scriptsize no}\tabularnewline
 &  &  & {\scriptsize $d(\mathbf{x},\mathbf{w})=\left\Vert \mathbf{x}-\mathbf{w}\right\Vert $} & \tabularnewline
\hline 
{\scriptsize SGNG} & {\scriptsize Margin Maximization} & {\scriptsize $\left\{ \epsilon,\mathbf{W}\right\} $} & {\scriptsize Euclidean Distance} & {\scriptsize no}\tabularnewline
 &  &  & {\scriptsize $d(\mathbf{x},\mathbf{w})=\left\Vert \mathbf{x}-\mathbf{w}\right\Vert $} & \tabularnewline
\hline 
{\scriptsize H2MLVQ} & {\scriptsize Margin Maximization} & {\scriptsize $\left\{ \epsilon,\mathbf{W}\right\} $} & {\scriptsize Harmonic to minimum distance} & {\scriptsize no}\tabularnewline
 &  &  &  & \tabularnewline
\hline 
{\scriptsize GRLVQ} & {\scriptsize Margin Maximization} & {\scriptsize $\left\{ \epsilon,\mathbf{W}\right\} $} & {\scriptsize Adaptive distance} & {\scriptsize $\sum_{i}\lambda_{i}=1$}\tabularnewline
 & {\scriptsize gradient descent} &  & {\scriptsize $d^{\lambda}(\mathbf{x},\mathbf{w})=\sum_{i}\lambda_{i}(x_{i}-w_{i})^{2}$} & \tabularnewline
\hline 
{\scriptsize GMLVQ} & {\scriptsize Margin Maximization} & {\scriptsize $\left\{ \epsilon,\Lambda,\mathbf{W}\right\} $} & {\scriptsize Adaptive distance} & {\scriptsize Matrix $\Lambda=\Omega\Omega^{T}$}\tabularnewline
 &  &  & {\scriptsize $d^{\Lambda}(\mathbf{x},\mathbf{w})=(\mathbf{x}-\mathbf{w})^{T}\Lambda(\mathbf{x}-\mathbf{w})$} & {\scriptsize with $\sum_{i}\Lambda_{ii}=1$}\tabularnewline
\hline 
{\scriptsize LGRLVQ} & {\scriptsize Margin Maximization} & {\scriptsize $\left\{ \epsilon,\mathbf{W}\right\} $} & {\scriptsize Adaptive distance} & {\scriptsize $\sum_{i}\lambda_{i}=1$}\tabularnewline
 & {\scriptsize gradient descent} &  & {\scriptsize $d^{\lambda}(\mathbf{x},\mathbf{w})=\sum_{i}\lambda_{i}(x_{i}-w_{i})^{2}$} & \tabularnewline
\hline 
{\scriptsize LGMLVQ} & {\scriptsize Margin Maximization} & {\scriptsize $\left\{ \epsilon,\Lambda,\mathbf{W}\right\} $} & {\scriptsize Adaptive distance} & {\scriptsize Matrix $\Lambda=\Omega\Omega^{T}$}\tabularnewline
 &  &  & {\scriptsize $d^{\Lambda}(\mathbf{x},\mathbf{w})=(\mathbf{x}-\mathbf{w})^{T}\Lambda(\mathbf{x}-\mathbf{w})$} & {\scriptsize with $\sum_{i}\Lambda_{ii}=1$}\tabularnewline
\hline 
{\scriptsize KRSLVQ} & {\scriptsize Likelihood Ratio Maximization} & {\scriptsize $\left\{ \epsilon,\sigma_{k},\sigma,\gamma\right\} $} & {\scriptsize Euclidean distance in feature space $\mathbf{F}$} & {\scriptsize no}\tabularnewline
 & {\scriptsize kernelized} &  & {\scriptsize $d(\Phi(\mathbf{x}),\mathbf{w}^{F})=\left\Vert \Phi(\mathbf{x})-\mathbf{w}^{F}\right\Vert $} & \tabularnewline
\hline 
\end{tabular}
\end{table}

Table 2 shows the parameter values used for the 11 LVQ classifiers
for the three datasets. The expression ``id'' stands for equal values
to those shown in the first column.

\begin{table}[t]
\caption{Summary of parameter values used for each LVQ classifier for the three
datasets.}

\centering{}%
\begin{tabular}{c|c|c|c}
\multicolumn{1}{c}{} & \multicolumn{1}{c}{{\scriptsize Multi-modal}} & \multicolumn{1}{c}{{\scriptsize Image Segmentation}} & {\scriptsize USPS}\tabularnewline
\hline 
\hline 
{\scriptsize LVQ2.1 } & {\scriptsize }%
\begin{tabular}{l}
{\scriptsize $\epsilon_{0}^{p}=0.05$}\tabularnewline
{\scriptsize $t_{0}^{p}=0$}\tabularnewline
{\scriptsize $s=0.01$}\tabularnewline
\end{tabular} & {\scriptsize id} & {\scriptsize id}\tabularnewline
\hline 
{\scriptsize GLVQ } & {\scriptsize }%
\begin{tabular}{l}
{\scriptsize $\epsilon_{0}^{p}=0.05$}\tabularnewline
{\scriptsize $t=0$}\tabularnewline
\end{tabular} & {\scriptsize id} & {\scriptsize id}\tabularnewline
\hline 
{\scriptsize RSLVQ } & {\scriptsize }%
\begin{tabular}{l}
{\scriptsize $\epsilon_{0}^{p}=0.05$}\tabularnewline
{\scriptsize $t_{0}^{p}=0$}\tabularnewline
{\scriptsize $\sigma_{opt}=1.9858$}\tabularnewline
\end{tabular} & {\scriptsize }%
\begin{tabular}{l}
{\scriptsize $\sigma_{opt}=0.01$}\tabularnewline
\end{tabular} & {\scriptsize }%
\begin{tabular}{l}
{\scriptsize $\sigma_{opt}=0.01$}\tabularnewline
\end{tabular}\tabularnewline
\hline 
{\scriptsize SNG } & {\scriptsize }%
\begin{tabular}{l}
{\scriptsize $\epsilon_{0}^{p}=0.05$}\tabularnewline
{\scriptsize $t_{0}^{p}=0$}\tabularnewline
\end{tabular} & {\scriptsize id} & {\scriptsize id}\tabularnewline
\hline 
{\scriptsize SGNG } & {\scriptsize }%
\begin{tabular}{l}
{\scriptsize $\epsilon_{0}^{p}=0.05$}\tabularnewline
{\scriptsize $t_{0}^{p}=0$}\tabularnewline
{\scriptsize $N_{p\_max}=45$}\tabularnewline
\end{tabular} & {\scriptsize }%
\begin{tabular}{l}
{\scriptsize $N_{p\_max}=10$}\tabularnewline
\end{tabular} & {\scriptsize }%
\begin{tabular}{l}
{\scriptsize $N_{p\_max}=30$}\tabularnewline
\end{tabular}\tabularnewline
\hline 
{\scriptsize H2MLVQ } & {\scriptsize }%
\begin{tabular}{l}
{\scriptsize $\epsilon_{0}^{p}=0.05$}\tabularnewline
{\scriptsize $t_{0}^{p}=0$}\tabularnewline
\end{tabular} & {\scriptsize id} & {\scriptsize id}\tabularnewline
\hline 
{\scriptsize GRLVQ } & {\scriptsize }%
\begin{tabular}{l}
{\scriptsize $\epsilon_{0}^{p}=0.05$}\tabularnewline
{\scriptsize $t_{0}^{p}=0$}\tabularnewline
{\scriptsize $\epsilon_{0}^{r}=5\cdot10^{-6}$}\tabularnewline
{\scriptsize $t_{0}^{r}=500$}\tabularnewline
\end{tabular} & {\scriptsize }%
\begin{tabular}{l}
{\scriptsize $t_{0}^{r}=100$}\tabularnewline
\end{tabular} & {\scriptsize }%
\begin{tabular}{l}
{\scriptsize $t_{0}^{r}=100$}\tabularnewline
\end{tabular}\tabularnewline
\hline 
{\scriptsize GMLVQ } & {\scriptsize }%
\begin{tabular}{l}
{\scriptsize $\epsilon_{0}^{p}=0.05$}\tabularnewline
{\scriptsize $t_{0}^{p}=0$}\tabularnewline
{\scriptsize $\epsilon_{0}^{d}=5\cdot10^{-5}$}\tabularnewline
{\scriptsize $t_{0}^{d}=500$}\tabularnewline
{\scriptsize $\epsilon_{0}^{m}=1\cdot10^{-6}$}\tabularnewline
{\scriptsize $t_{0}^{m}=500$}\tabularnewline
\end{tabular} & {\scriptsize }%
\begin{tabular}{l}
{\scriptsize $t_{0}^{d}=100$}\tabularnewline
{\scriptsize $t_{0}^{m}=100$}\tabularnewline
\end{tabular} & {\scriptsize }%
\begin{tabular}{l}
{\scriptsize $t_{0}^{d}=100$}\tabularnewline
{\scriptsize $t_{0}^{m}=100$}\tabularnewline
\end{tabular}\tabularnewline
\hline 
{\scriptsize LGRLVQ } & {\scriptsize }%
\begin{tabular}{l}
{\scriptsize $\epsilon_{0}^{p}=0.05$}\tabularnewline
{\scriptsize $t_{0}^{p}=0$}\tabularnewline
{\scriptsize $\epsilon_{0}^{d}=5\cdot10^{-5}$}\tabularnewline
{\scriptsize $t_{0}^{d}=100$}\tabularnewline
\end{tabular} & {\scriptsize id} & {\scriptsize id}\tabularnewline
\hline 
{\scriptsize LGMLVQ } & {\scriptsize }%
\begin{tabular}{l}
{\scriptsize $\epsilon_{0}^{p}=0.05$}\tabularnewline
{\scriptsize $t_{0}^{p}=0$}\tabularnewline
{\scriptsize $\epsilon_{0}^{d}=1\cdot10^{-3}$}\tabularnewline
{\scriptsize $t_{0}^{d}=100$}\tabularnewline
{\scriptsize $\epsilon_{0}^{m}=5\cdot10^{-5}$}\tabularnewline
{\scriptsize $t_{0}^{m}=100$}\tabularnewline
\end{tabular} & {\scriptsize id} & {\scriptsize id}\tabularnewline
\hline 
{\scriptsize KRSLVQ } & {\scriptsize }%
\begin{tabular}{l}
{\scriptsize $\epsilon_{0}^{p}=0.05$}\tabularnewline
{\scriptsize $t_{0}^{p}=0$}\tabularnewline
{\scriptsize $\sigma_{opt}=1$}\tabularnewline
\end{tabular} & {\scriptsize }%
\begin{tabular}{l}
{\scriptsize $\sigma_{opt}=0.01$}\tabularnewline
\end{tabular} & {\scriptsize }%
\begin{tabular}{l}
{\scriptsize $\sigma_{opt}=0.5$}\tabularnewline
\end{tabular}\tabularnewline
\hline 
\end{tabular}
\end{table}

\subsection{Multi-modal dataset}

Table \ref{tab:Classification-Error} (the second column) shows the
results obtained for the multi-modal dataset using 10-fold cross validation.
The initialization of prototypes was random in the mean of the whole
dataset. This setting allows quantifying the sensitivity to the initial
conditions of each algorithm under study. The number of prototypes
per class was set to $N_{p}=15$. As shown in Table \ref{tab:Classification-Error}
(second column) the algorithms that update more than one prototype
per iteration such as H2M-LVQ, SNG and SGNG, obtained the best results
together with GLVQ. A multi-comparison statistical test was performed
which showed that H2M-LVQ is significantly better than 6 other algorithms
(LVQ 2.1, RSLVQ, GMLVQ, LGRLVQ, LGMLVQ, KRSLVQ) with a 95\% confidence
interval of the mean difference. The statistical test indicates that
there are 4 algorithms that are not significantly different from H2M-LVQ:
SGNG, SNG, GLVQ and GRLVQ. However, if we take any of these 4 algorithms
as a reference, the statistical test shows that they are statistically
significantly different from only 3 LVQ classifiers instead of 6 as
was done by H2MLVQ. The results in Table \ref{tab:Classification-Error}
show that the performance of LVQ 2.1 is inferior to all other algorithms.
This is because LVQ 2.1 does not have a associated functional and
it is very sensitive to the initial condition of the prototypes. In
practice LVQ 2.1 needs a pre-processing technique for finding a good
initial position of the prototypes. For methods using the adaptive
(local) metric, such as: GRLVQ, LGRLVQ, GMLVQ and LGMLVQ, the classification
error is higher that obtained by GLVQ because these methods are similar
to GLVQ in the early iterations, and are prone to over-fitting the
adaptive metric while trying to find a good performance.

\begin{table}[t]
\caption{Average Classification Errors obtained by using 10-fold Cross Validation.
(The standard deviation is shown within brackets.)\label{tab:Classification-Error}}

\centering{}%
\begin{tabular}{c|c|c|c|c}
\multicolumn{1}{c}{} & \multicolumn{1}{c}{{\scriptsize Multi-modal}} & \multicolumn{1}{c}{{\scriptsize Image Segmentation}} & \multicolumn{1}{c}{{\scriptsize USPS{*}}} & {\scriptsize USPS}\tabularnewline
\hline 
\hline 
{\scriptsize LVQ2.1 } & {\scriptsize 0.3289 (0.0400)} & {\scriptsize 0.2886 (0.0387)} & {\scriptsize 0.2390 (0.0361)} & {\scriptsize 0.2700 (0.0954)}\tabularnewline
\hline 
{\scriptsize GLVQ } & {\scriptsize 0.0669 (0.0141)} & {\scriptsize 0.1205 (0.1497)} & {\scriptsize 0.0570 (0.0173)} & {\scriptsize 0.0831 (0.0036)}\tabularnewline
\hline 
{\scriptsize RSLVQ } & {\scriptsize 0.1583 (0.0346)} & {\scriptsize 0.2124 (0.0539)} & \textbf{\scriptsize 0.0415 (0.0076)} & {\scriptsize 0.0566 (0.0141)}\tabularnewline
\hline 
{\scriptsize SNG } & {\scriptsize 0.0678 (0.0141)} & {\scriptsize 0.1205 (0.0300)} & {\scriptsize 0.0570 (0.0141)} & \textbf{\scriptsize 0.0410 (0.0714)}\tabularnewline
\hline 
{\scriptsize SGNG } & {\scriptsize 0.0732 (0.1873)} & {\scriptsize 0.2200 (0.2121)} & {\scriptsize 0.0815 (0.2141)} & {\scriptsize 0.0922 (0.2213)}\tabularnewline
\hline 
{\scriptsize H2MLVQ } & \textbf{\scriptsize 0.0294 (0.0141)} & {\scriptsize 0.1743 (0.0387)} & {\scriptsize 0.0530 (0.0632)} & {\scriptsize 0.0455 (0.0917)}\tabularnewline
\hline 
{\scriptsize GRLVQ } & {\scriptsize 0.1183 (0.0224)} & {\scriptsize 0.0881 (0.0141)} & {\scriptsize 0.0970 (0.0245)} & {\scriptsize 0.0976 (0.1497)}\tabularnewline
\hline 
{\scriptsize GMLVQ } & {\scriptsize 0.1142 (0.0141)}\emph{\scriptsize{} } & {\scriptsize 0.0890 (0.0300)} & {\scriptsize 0.1165 (0.0224)} & {\scriptsize 0.1321 (0.0837)}\tabularnewline
\hline 
{\scriptsize LGRLVQ } & {\scriptsize 0.1031 (0.0361)}\emph{\scriptsize{} } & {\scriptsize 0.0531 (0.0224)} & {\scriptsize 0.1040 (0.0632)} & {\scriptsize 0.1091(0.0200)}\tabularnewline
\hline 
{\scriptsize LGMLVQ } & {\scriptsize 0.0955 (0.0458)}\emph{\scriptsize{} } & \textbf{\scriptsize 0.0357 (0.0224)} & {\scriptsize 0.1012 (0.0265)} & {\scriptsize 0.1074 (0.0224)}\tabularnewline
\hline 
{\scriptsize KRSLVQ } & {\scriptsize 0.1133 (0.0387) } & {\scriptsize 0.1587 (0.0332)} & {\scriptsize 0.0448 (0.0346)} & {\scriptsize 0.0514 (0.0224)}\tabularnewline
\hline 
\end{tabular}
\end{table}

\subsection{Image Segmentation dataset}

In this experiment the prototypes were initialized in the mean of
each class to avoid initialization sensitivity. The number of prototypes
per class was set to $N_{p}=1$. The algorithms based on distance
learning (matrix or relevance learning) reached a better performance,
as shown in the third column of Table \ref{tab:Classification-Error}.
The mean (std) using 10-fold cross validation are shown in this table.
The lowest classification error was obtained by LGMLVQ with a mean
value of 0.0357. This is significantly better than 6 other algorithms
(KRSLVQ, RSLVQ, LGRLVQ, SGNG, H2M-LVQ and LVQ) with a 95\% confidence
interval of the mean difference according to the multi-comparison
statistical test using Bonferroni adjustment. On the other hand, LGMLVQ
is not significantly different from GLVQ, GRLVQ, SNG and GMLVQ. But,
as discussed in the previous section, the means of these 4 algorithms
are significantly different from only three algorithms instead of
6 as was done by LGMLVQ. Between GMLVQ and GRLVQ there is no statistically
significant difference, with a 95\% confidence interval of the mean
difference. In the case of local distance learning (LGMLVQ and LGRLVQ),
the matrix method obtained better performance than the relevance version,
which is significantly different with a confidence interval of 95\%.
The methods based on distance learning obtained better performance
due to their capacity of modifying the shape of the receptive field
of prototypes, locally or globally, being robust against a dataset
with heterogeneous features such as the Image Segmentation dataset.

\subsection{USPS dataset}

For both USPS and USPS{*} datasets the prototypes were initialized
in the mean of each class and the number of prototypes per class was
set to $N_{p}=3$. The methods based on likelihood ratio maximization,
RSLVQ and KRSLVQ, obtained higher performance than the methods based
on margin maximization. For USPS{*} both RSLVQ and KRSLVQ reached
average classification errors that are significantly better than the
other 6 algorithms with a 95\% confidence interval. The best performance
was obtained by RSLVQ with a mean of 0.0415. The multi-comparison
statistical test indicates that the means obtained by RSLVQ and KRSLVQ
are not significantly different with a 95\% confidence interval. The
algorithms based on likelihood ratio maximization achieved higher
performance in this dataset because the best prototypes are not necessarily
located in the centroids of each cluster, which allows obtaining better
performance in datasets that are very overlapped. In the case of USPS
the best performance was obtained by SNG followed by H2M-LVQ, but
the methods based on likelihood-ratio maximization RSLVQ and KRSLVQ
appear closely in the third and fourth places as is shown in Table
\ref{tab:Classification-Error} (the fifth column).

\section{Open Problems}

In this section some open problems and challenges in the field of
LVQ classifiers are presented.
\begin{enumerate}
\item \emph{Principled Approach to LVQ.} Although GLVQ provides a cost function
and it can be shown to be a margin maximizer method, the cost function
is not derived from first principles, such as probabilistic or information
theory.
\item \emph{Sparsity}. The real world datasets are becoming larger and larger,
and the computational cost of applying a prototype-based method, such
as the LVQ classifier, keeps growing. For this reason, recently several
sparsity approaches have been extended to LVQ classifiers which allow
obtaining a linear training time without losing classification performance
\cite{williams2001using,hofmann2013efficient}.
\item \emph{Semi-supervised learning}. In the real world due to the increasing
size of datasets it is no longer possible to put labels on all samples.
For this reason, it is necessary to adapt the LVQ classifiers to semi-supervised
learning or an active learning framework \cite{chapelle2006semi}.
\item \emph{Visualization} \emph{of LVQ classifiers}. Prototype-based methods
such as LVQ classifiers allow the prototypical representation of the
data in the input space. This is an advantage because when the prototypes
and data can be visualized, the classifier can be interpreted easily.
But when the LVQ classifiers work in a space different from the data
space, such as in kernelized and relational variants, the classifiers
are no longer easily visualized and interpreted. Not losing this natural
capacity of the early LVQ classifiers is of interest \cite{nova2013online,hammer2013visualize}.
\item \emph{Active learning.} For improving the generalization ability of
the prototype-based methods such as LVQ classifiers, active learning
can be used. This method gives the learner the capability of selecting
samples during training. Furthermore, using the active learning approach,
the generalization ability of the model can be increased as well as
its learning speed \cite{Schleif2007,baum1991neural,mitra2004probabilistic}.
\end{enumerate}

\section{Conclusions}

We have presented a review of the most relevant LVQ classifiers developed
in the last 25 years. We introduced a taxonomy of LVQ classifiers
as a general framework. Two different main cost functions have been
proposed in the literature: margin maximization and likelihood ratio
maximization. LVQ classifiers based on margin maximization have been
demonstrated to have good performance in many problems. These methods
put the prototypes in the centroid of the data which makes them less
flexible with overlapping data. LVQ classifiers based on a likelihood
ratio maximization are an alternative that uses a probabilistic approach,
in which the prototypes are not put in the centroid of the classes,
which gives them flexibility in the case of overlapping data. On the
other hand, the LVQ classifiers based on an adaptive metric have reached
the best performance in heterogeneous feature datasets. Also, LVQ
classifiers which update more than one prototype per iteration are
less sensitive to initial conditions and get better performance in
multi-modal datasets.

Recently, LVQ classifiers such as kernelized or relational LVQ classifiers
have been based on data representation for improving the performance
of the classifiers when the data is more complex. With relational
LVQ classifiers there is a more general representation of the data
which allows working with non-Euclidean spaces. In this sense, the
recent approaches based on dis-/similarities capture the inherent
data structure naturally which should improve the performance of the
classifier. The experiments done in this work have shown that there
is no free lunch; each method has its own pros and cons. The different
LVQ methods were designed for dealing with specific problems and datasets.
From a more general point of view, LVQ classifiers have been demonstrated
to be very competitive classifiers, and further research is needed
to achieve the greatest success in pattern recognition tasks.
\begin{acknowledgements}
This work was funded by CONICYT-CHILE under grant FONDECYT 1110701.
\end{acknowledgements}

\end{document}